\documentclass{article}
\usepackage{ijcai19}

\usepackage{times}
\usepackage{soul}
\usepackage{latexsym}

\usepackage{enumerate}
\usepackage{graphicx}
\usepackage{subfigure}
\usepackage{multirow}

\usepackage{framed,multirow}
\usepackage{times}
\usepackage{graphicx} 
\usepackage{subfigure}
\usepackage{amssymb}
\usepackage{latexsym}

\usepackage{algorithm}
\usepackage{algorithmic}

\usepackage{hyperref}

\usepackage{url}
\usepackage{xcolor}
\usepackage{amsmath}
\usepackage{amsxtra}
\usepackage{braket}
\usepackage{array}
\usepackage{color}
\usepackage{url}
\usepackage{placeins}
\usepackage{threeparttable}

\usepackage{times}
\usepackage{balance}
\usepackage{enumerate}
\usepackage{multirow}
\usepackage{pifont}
\usepackage{marvosym}
\usepackage{ifsym}
\title{Learning Vertex Convolutional Networks for Graph Classification}

\author{
Lu Bai$^1$
\and
Lixin Cui$^1$\and
Shu Wu$^{2}$\and
Yuhang Jiao$^{1}$\And
Edwin R. Hancock$^3$
\affiliations
$^1$Central University of Finance and Economics, Beijing, China\\
$^2$Institute of Automation, Chinese Academy of Sciences, Beijing, China\\
$^3$University of York, York, UK\\
}

\begin{document}

\maketitle

\begin{abstract}
In this paper, we develop a new aligned vertex convolutional network model to learn multi-scale local-level vertex features for graph classification. Our idea is to transform the graphs of arbitrary sizes into fixed-sized aligned vertex grid structures, and define a new vertex convolution operation by adopting a set of fixed-sized one-dimensional convolution filters on the grid structure. We show that the proposed model not only integrates the precise structural correspondence information between graphs but also minimises the loss of structural information residing on local-level vertices. Experiments on standard graph datasets demonstrate the effectiveness of the proposed model.
\end{abstract}

\section{Introduction}\label{s1}

Graph-based representations are powerful tools to analyze real-world structured data that encapsulates pairwise relationships between its parts~\cite{DBLP:conf/nips/DefferrardBV16,DBLP:journals/tnn/ZambonAL18}. One fundamental challenge arising in the analysis of graph-based data is to represent discrete graph structures as numeric features that preserve the topological information. Due to the recent successes of deep learning networks in computer vision problems, many researchers have devoted their efforts to generalizing Convolutional Neural Networks (CNNs)~\cite{DBLP:conf/cvpr/VinyalsTBE15,DBLP:journals/cacm/KrizhevskySH17} to the graph domain. These neural networks on graphs are now widely known as Graph Convolutional Networks (GCNs)~\cite{DBLP:journals/corr/KipfW16}, and
have proven to be an effective way to extract highly meaningful statistical features for graph classification problems~\cite{DBLP:conf/nips/DefferrardBV16}.

Generally speaking, most existing state-of-the-art graph convolutional networks are developed based on two strategies, i.e., a) the spectral and b) the spatial strategies. Specifically, approaches based on the spectral strategy employ the property of the convolution operator from the graph Fourier domain that is related to spectral graph theory~\cite{DBLP:journals/corr/BrunaZSL13}. By transforming the graph into the spectral domain through the eigenvectors of the Laplacian matrix, these methods perform the filter operation by multiplying the graph by a series of filter coefficients~\cite{DBLP:journals/corr/BrunaZSL13,DBLP:conf/nips/RippelSA15,DBLP:journals/corr/HenaffBL15}. Unfortunately, most spectral-based approaches demand the size of the graph structures to be the same and cannot be performed on graphs with different sizes and Fourier bases. As a result, approaches based on the spectral strategy are usually applied to vertex classification tasks. Methods based on the spatial strategy, on the other hand, generalize the convolution operation to the spatial structure of a graph by propagating features between neighboring vertices~\cite{DBLP:journals/corr/VialatteGM16,DBLP:conf/nips/DuvenaudMABHAA15,DBLP:conf/nips/AtwoodT16}. Since spatial-based approaches are not restricted to the same graph structure, these methods can be directly applied to graph classification problems. Unfortunately, most existing spatial-based methods have relatively poor performance on graph classifications. The reason for this ineffectiveness is that these methods tend to directly sum up the extracted local-level vertex features from the convolution operation as global-level graph features through a SumPooling layer. As a result, the local topological information residing on the vertices of a graph may be discarded.

To address the shortcoming of the graph convolutional networks associated with SumPooling, a number of methods focusing on local-level vertex information have been proposed. For instance, \cite{DBLP:conf/icml/NiepertAK16} have developed a different graph convolutional network by re-ordering the vertices and converting each graph into fixed-sized vertex grid structures, where standard one-dimensional CNNs can be directly used. \cite{DBLP:conf/aaai/ZhangCNC18} have developed a novel Deep Graph Convolutional Neural Network model to preserve more vertex information through global graph topologies. Specifically, they propose a new SortPooling layer to transform the extracted vertex features of unordered vertices from the spatial graph convolution layers into a fixed-sized vertex grid structure. Then a traditional convolutional operation can be performed by sliding a fixed-sized filter over the vertex grid structures to further learn the topological information. The aforementioned methods focus more on local-level vertex features and outperform state-of-the-art graph convolutional network models on graph classification tasks. However, they tend to sort the vertex order based on the local structure descriptor of each individual graph. As a result, they cannot easily reflect the accurate topological correspondence information between graph structures. Furthermore, these approaches also lead to significant information loss. This usually occurs when they form a fixed-sized vertex grid structure and some vertices associated with lower ranking may be discarded. In summary, developing effective methods to preserve the structural information residing in graphs still remains a significant challenge.

To overcome the shortcoming of the aforementioned methods, we propose a new graph convolutional network model, namely the Aligned Vertex Convolutional Network, to learn multi-scale features from local-level vertices for graph classification. One key innovation of the proposed model is the identification of the transitively aligned vertices between graphs. That is, given three vertices $v$, $w$ and $x$ from three sample graphs, assume $v$ and $x$ are aligned, and $w$ and $x$ are aligned, the proposed model can guarantee that $v$ and $w$ are also aligned. More specifically, the new model utilizes the transitive alignment procedure to transform different graphs into fixed-sized aligned vertex grid structures with consistent vertex orders. Overall, the main contributions are threefold.

\textbf{First}, we propose a new vertex matching method to transitively align the vertices of graphs. We show that this matching procedure can establish reliable vertex correspondence information between graphs, by gradually minimizing the inner-vertex-cluster sum of squares over the vertices of all graphs through a $k$-means clustering method.

\textbf{Second}, with the transitive alignment information over a family of graphs to hand, we show how the graphs of arbitrary sizes can be mapped into fixed-sized aligned vertex grid structures. The resulting Aligned Vertex Convolutional Network model is defined by adopting fixed-sized one-dimensional convolution filters on the grid structure to slide over the entire ordered aligned vertices. We show that the proposed model can effectively learn the multi-scale characteristics residing on the local-level vertex features for graph classifications. Moreover, since all the original vertex information will be mapped into the aligned vertex grid structure through the transitive alignment, the grid structure not only precisely integrates the structural correspondence information but also minimises the loss of structural information residing on local-level vertices. As a result, the proposed model addresses the shortcomings of information loss and imprecise information representation arising in existing graph convolutional networks associated with SortPooling or SumPooling.

\textbf{Third}, we empirically evaluate the performance of the proposed model on graph classification problems. Experiments on benchmark graph datasets demonstrate the effectiveness.

\section{Transitive Vertex Alignment Method}\label{s2}

One main objective of this work is to convert graphs of arbitrary sizes into the fixed-sized aligned vertex grid structures, so that a fixed-sized convolution filter can directly slide over the grid structures to learn local-level structural features through vertices. To this end, we need to identify the correspondence information between graphs.

In this section, we introduce a new matching method to transitively align the vertices. We commence by designating a family of prototype representations that encapsulate the principle characteristics over all vectorial vertex representations in a set of graphs $\mathbf{G}$. Assume there are $n$ vertices from all graphs in $\mathbf{G}$, and the associated $K$-dimensional vectorial representations of these vertices are $\mathbf{{R}}^K =\{\mathrm{R}_1^K,\mathrm{R}_2^K,\ldots,\mathrm{R}_n^K\}$. We employ $k$-means~\cite{witten2011data} to locate $M$ centroids over all representations in $\mathbf{{R}}^K$. Specifically, given $M$ clusters $\Omega=(c_1,c_2,\ldots,c_M)$, the aim of $k$-means is to minimize the objective function
\begin{equation}
\arg\min_{\Omega}  \sum_{j=1}^M \sum_{\mathrm{R}_i^K \in c_j} \|\mathrm{R}_i^K- \mu_j^K\|^2,\label{kmeans}
\end{equation}
where $\mu_j^K$ is the mean of the vertex representations belonging to the $j$-th cluster $c_j$. Since Eq.(\ref{kmeans}) minimizes the sum of the square Euclidean distances between the vertex points $\mathrm{R}_i^K$ and the centroid point of cluster $c_j$, the set of $M$ centroid points $\mathbf{PR}^K=\{\mu_1^K,\cdots,\mu_j^K,\cdots,\mu_M^K\}$ can be seen as a family of $K$-dimensional \textbf{prototype representations} that encapsulate representative characteristics over all graphs in $\mathbf{G}$.

To establish the correspondence information between the graph vertices over all graphs in $\mathbf{G}$, we align the vectorial vertex representations of each graph to the prototype representations in $\mathbf{PR}^K$. Our alignment is similar to that introduced in~\cite{DBLP:conf/icml/Bai0ZH15} for point matching in a pattern space. Specifically, for each sample graph $G_p(V_p,E_p)\in {\mathbf{G}}$ and the associated $K$-dimensional vectorial representation of each vertex $v_i\in V_p$, we compute a $K$-level affinity matrix in terms of the Euclidean distances between the two sets of points as
\begin{align}
A^K_p(i,j)=\|\mathrm{R}_i^K - \mu_j^K\|_2.\label{AffinityM}
\end{align}
where $A^K_p$ is a ${|V_p|}\times {M}$ matrix, and each element $R^K_p(i,j)$ represents the distance between the vectorial representation $\mathrm{{R}}_{p;i}^K$ of vertex $v_\in V_p$ and the $j$-prototype representation $\mu_j^K\in \mathbf{PR}^K$. If the value of $A^K_p(i,j)$ is the smallest in row $i$, the vertex $v_i$ is aligned to the $j$-th prototype representation. Note that for each graph there may be two or more vertices aligned to the same prototype representation. We record the correspondence information using the $K$-level correspondence matrix
$C^K_p\in \{0,1\}^{|V_p|\times M}$
\begin{equation}
C^K_p(i,j)=\left\{
\begin{array}{cl}
1   & \small{\mathrm{if} \  A^K_p(i,j) \ \mathrm{is \ the \ smallest \ in \ row } \ i} \\
0   & \small{\mathrm{otherwise}}.
\end{array} \right.
\label{CoMatrix}
\end{equation}

For a pair of graphs $G_p$ and $G_q$, if their vertices $v_p$ and $v_q$ are aligned to the same prototype representation, we say that $v_p$ and $v_q$ possess similar characteristics and are also aligned. Thus, we can identify the transitive alignment information between the vertices of all graphs in $\mathbf{G}$, by aligning their vertices to the same set of prototype representations. The alignment process is equivalent to assigning the vectorial representation $\mathrm{{R}}_{p;i}^K$ of each vertex $v_i\in V_p$ to the mean $\mu_i^K$ of the cluster $c_i^K$. Thus, the proposed alignment procedure can be seen as an optimization process that gradually minimizes the inner-vertex-cluster sum of squares over the vertices of all graphs through $k$-means, and can establish reliable vertex correspondence information over all graphs.

\section{Learning Vertex Convolutional Networks}\label{s3}

In this section, we develop a new vertex convolutional network model for graph classification. Our idea is to employ the transitive alignment information over a family of graphs and convert the arbitrary sized graphs into fixed-sized aligned vertex grid structures. We then define a vertex convolution operation by adopting a set of fixed-sized one-dimensional convolution filters on the grid structure. With the new vertex convolution operation to hand, the proposed model can extract the original aligned vertex grid structure as a new grid structure with a reduced number of packed aligned vertices, i.e., the extracted multi-scale vertex features learned through the convolutional operation is packed into the new grid structure. Finally, we employ the Softmax layer to read the extracted vertex features and predict the graph class.

\subsection{Aligned Vertex Grid Structures of Graphs}
In this subsection, we show how to convert graphs of different sizes into fixed-sized aligned vertex grid structures. For each sample graph $G_p(V_p,E_p)$ from the graph set $\mathbf{G}$ defined earlier, assume each of its vertices $v_p\in V_p$ is represented as a $c$-dimensional feature vector. Then the features of all the $n$ ($n=|V_p|$) vertices can be encoded using the $n\times c$ matrix $F_p$ (i.e., $F_p\in \mathbb{R}^{n\times c}$). If $G_p$ are vertex attributed graphs, $F_p$ can be the one-hot encoding matrix of the vertex labels. For un-attributed graphs, we propose to use the vertex degree as the vertex label. Based on the transitive alignment method defined in Section~\ref{s2}, we commence by identifying the family of the $K$-dimensional prototype representations in $\mathbf{PR}^K=\{\mu_1^K,\ldots,\mu_j^K,\ldots,\mu_M^K \}$ of $\mathbf{G}$. For each graph $G_p\in \mathbf{G}$, we compute the $K$-level vertex correspondence matrix $C^K_p$, where the row and column of $C^K_p$ are indexed by the vertices in $V_p$ and the prototype representations in $\mathbf{PR}^K$, respectively. With $C^K_p$ to hand, we compute the $K$-level aligned vertex feature matrix for $G_p$ as
\begin{equation}
{X}_{p}^{K}= (C^K_p)^T F_p,\label{alignDB}
\end{equation}
where ${X}_{p}^{K}$ is a $M\times c$ matrix and each row of ${X}_{p}^{K}$ represents the feature of a corresponding aligned vertex. Since ${X}_{p}^{K}$ is computed by mapping the original feature information of each vertex $v_p\in V_p$ to that of the new aligned vertices indexed by the corresponding prototypes in $\mathbf{PR}^K$, it encapsulates all the original vertex feature information of $G_p$.

For constructing the fixed-sized aligned vertex grid structure for each graph $G_p\in \mathbf{G}$, we need to establish a consistent vertex order for all graphs in $\mathbf{G}$. As the vertices are all aligned to the same prototype representations, the vertex orders can be determined by reordering the prototype representations. To this end, we construct a prototype graph that captures the pairwise similarity between the prototype representations, then we reorder the prototype representations based on their degree. This process is equivalent to sorting the prototypes in order of average similarity to the remaining ones. Specifically, for the $K$-dimensional prototype representations in $\mathbf{PR}^K$, we compute the prototype graph as $G_{\mathrm{R}}(V_{\mathrm{R}},E_{\mathrm{R}})$, where each vertex $v_j\in V_{\mathrm{R}}$ represents the prototype representation $\mu_j^K\in \mathbf{PR}^K$ and each edge $(v_j,v_k)\in E_{\mathrm{R}} $ represents the similarity between a pair of prototype representations $\mu_j^K\in \mathbf{PR}^K$ and $\mu_k^K\in \mathbf{PR}^K$. The similarity between two vertices of $G_{\mathrm{R}}$ is computed as \begin{equation}
s(\mu_j^K,\mu_k^K)=\exp (-\frac{\| \mu_j^K-\mu_k^K   \|_2}{K}).
\end{equation}
The degree of each prototype representation $\mu_j^K$ is $D_R(\mu_j^K)=\sum_{k=1}^{M}s(\mu_j^K,\mu_k^K)$. We sort the $K$-dimensional prototype representations in $\mathbf{PR}^K$ according to their degree $D_R(\mu_j^K)$. Then, we rearrange ${X}_{p}^{K}$ accordingly.

Finally, note that, to construct reliable grid structures for graphs, we employ the depth-based (DB) representations as the vectorial vertex representations to compute the required $K$-level vertex correspondence matrix $C_p^K$. The DB representation of each vertex is defined by measuring the entropies on a family of $k$-layer expansion subgraphs rooted at the vertex~\cite{DBLP:journals/pr/BaiH14}, where the parameter $k$ varies from $1$ to $K$. It is shown that such a $K$-dimensional DB representation encapsulates rich entropy content flow from each local vertex to the global graph structure, as a function of depth. The process of computing the correspondence matrix $C_p^K$ associated with DB representations is shown in the appendix file. When we vary the largest layer $K$ of the expansion subgraphs from $1$ to $L$ (i.e., $K\leq L$), we compute the final \textbf{aligned vertex grid structure} for each graph $G_p$ as
\begin{equation}
{X}_{p}= \sum_{K=1}^L \frac{{X}_{p}^{K}}{L},\label{AlignV}
\end{equation}
where ${X}_{p}$ is also a $M\times c$ matrix as same as ${X}_{p}^{K}$. Clearly, Eq.(\ref{AlignV}) transforms the original graphs $G_p\in \mathbf{G}$ of arbitrary sizes into a new aligned vertex grid structure with the same vertex number. Moreover, note that, the aligned vertex grid structure ${X}_{p}$ also preserve the original vertex feature information through the $K$-level aligned vertex feature matrix ${X}_{p}^{K}$.

\subsection{The Aligned Vertex Convolutional Network}


In this subsection, we develop a new Aligned Vertex Convolutional Network model that learns local-level vertex features for graph classifications. This model is defined by adopting a set of fixed-sized one-dimensional convolution filters on the aligned vertex grid structures and sliding the filter over the ordered aligned vertices to learn features, in a manner analogous to the standard convolution operation. Specifically, for each graph $G(V,E)\in \mathbf{G}$ and its associated aligned vertex grid structure ${X}\in \mathbb{R}^{M\times c}$ (\textbf{i.e., $M$ aligned vertices each with $c$ feature channels}), we denote the element of $X$ in the $e$-th row and $s$-th column as ${X}_{e,s}$, i.e., the $s$-th feature channel of the $e$-th aligned vertex. We pass ${X}$ to the convolution layer. Assume the size of the receptive field is $m$, i.e., the size of the one-dimensional convolution filter is $m$, the vertex convolution operation associated with $1$-stride takes the form
\begin{equation}
Z_{e,h}=\sigma(\sum_{s=1}^{c}(\sum_{j=1}^{m}W_{j}^{h,s}{X}_{e+j-1,s})+b^{h}),\label{V_convolution}
\end{equation}
where $Z_{e,h}$ is the element in the $e$-th row and $h$-th column of the new grid structure $Z$ after the convolution operation, the parameter $e$ satisfies $e\leq M-m+1$, $W_{j}^{h,s}$ is the $j$-th element of the convolution filter that maps the $s$-th feature channel of $X$ to the $h$-th feature channel of $Z$, $b^h$ is the bias of the $h$-th convolution filter, and $\sigma$ is the activation function.

\begin{figure}
 \vspace{-0pt}
 \centering
\includegraphics[width=0.80\linewidth]{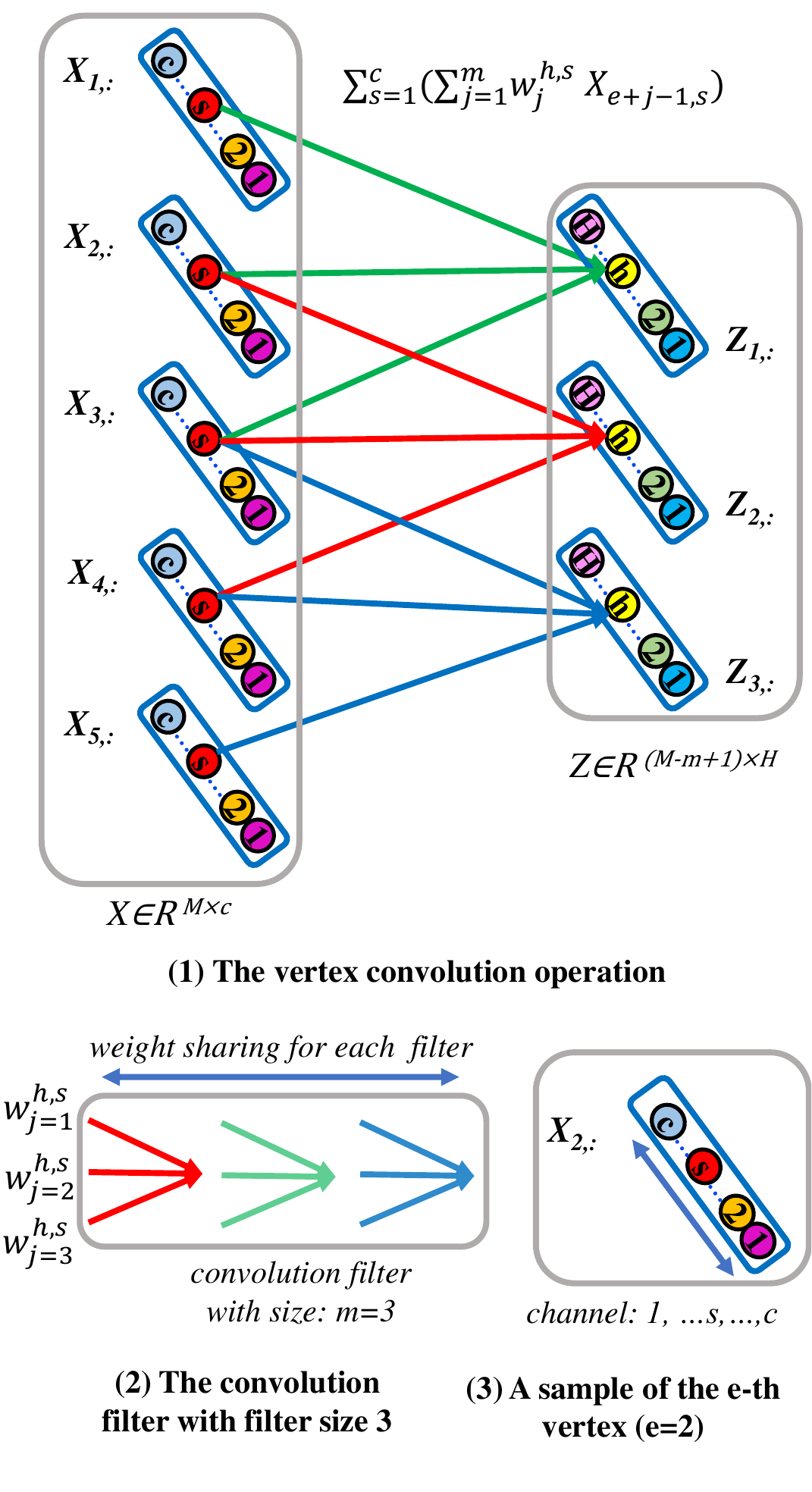}
 \vspace{-20pt}
  \caption{The procedure of the vertex convolution.}\label{f:vconv}
 \vspace{-20pt}
\end{figure}

An example of the vertex convolution operation defined by Eq.(\ref{V_convolution}) are show in Figure~\ref{f:vconv}. The vertex convolution operation consists of two computational steps. In the first step, the convolution filter $\sum_{s=1}^{c}(\sum_{j=1}^{m}W_{j}^{h,s}{X}_{e+j-1,s})$ is applied to map the $e$-th aligned vertex $X_{e,:}$ as well as its neighbor vertices $X_{e+j-1,:}$ ($j=2,3$) into a new feature value, associated with all the $c$ feature channels of these vertices. Specifically, Figure~\ref{f:vconv}.(1) exhibits this process. Here, assume the vertex index $e=2$, the convolution filter size $m=3$, and we focus on the $2$-nd aligned vertex $X_{2,:}$ of ${X}\in \mathbb{R}^{M\times c}$. The convolution filter $\sum_{s=1}^{c}(\sum_{j=1}^{m}W_{j}^{h,s}{X}_{2+j-1,s})$ represented by the red lines first maps the $s$-th feature channels of the $2$-nd aligned vertex $X_{2,:}$ as well as its neighbor vertices $X_{3,:}$ and $X_{4,:}$ into a new single value by $\sum_{j=1}^{m}W_{j}^{h,s}{X}_{2+j-1,s}$, and then sums up the values computed through all the $c$ channels as the $h$-th feature channel of $Z_{2,:}$. Moreover, we need to slide the convolution filter over all the aligned vertices, and this requires three convolution filters represented by the green, red and blue lines respectively. The weights for the three filters are shared, i.e., they are in fact the same filter. Finally, the second step $\sigma(\mathcal{X}_h +b^h)$, where $\mathcal{X}_h:=\sum_{s=1}^{c}(\sum_{j=1}^{m}W_{j}^{h,s}{X}_{e+j-1,s})$, applies the
Relu function associated with the bias $b^h$ and outputs the final result as $Z_{e,h}$.

To further extract the multi-scale features of a graph associated with its aligned vertex grid structure ${X}\in \mathbb{R}^{M\times c}$, we stack multiple vertex convolution layers defined as follows
\begin{equation}
Z_{e,h}^t=\sigma(\sum_{s=1}^{c}(\sum_{j=1}^{m} W_{j}^{t,h,s}{X}_{e+j-1,s}^{t-1})+b^{t,h}),\label{vcon_operation}
\end{equation}
where $X^0$ is the input aligned vertex grid structure $X$, and the corresponding notations of the symbols are listed in Table~\ref{T:notation}. After a number of vertex convolution operations, we employ the Softmax layer to read the extracted features computed from the vertex convolution layers and predict the graph class for graph classifications.

\begin{table}
\vspace{-0pt}
\centering {
\footnotesize
\caption{Important Terms and Notations}\label{T:notation}
\vspace{0pt}

\begin{tabular}{|c|c|}

  \hline
 ~Symbol ~      ~ &~Defitions~                                                         \\ \hline \hline
 ~node $e$~       &~ $\textrm{the e-th vertex}$  ~                                     \\ \hline

 ~$Z_{e,h}^{t}$~  &~ $\textrm{the h-th feature channel of vertex (e) in layer t}$ ~    \\ \hline

 ~$W^{t,h,s}$~    &~ $\textrm{the filter that maps to the h-th feature channel in}$ ~    \\
 ~$ $~            &~ $\textrm{layer t from the s-th feature channel in layer t-1 }$ ~   \\ \hline

 ~$W_j^{t,h,s}$~  &~ $\textrm{the j-th element of the filter that maps to the h-th }$ ~    \\
 ~$ $~            &~ $\textrm{feature channel in layer t from the s-th feature }$ ~   \\
 ~$ $~            &~ $\textrm{channel in layer t-1}$ ~   \\  \hline

 ~$b^{t,h}$~      &~ $\textrm{the bias of the h-th filter in layer t}$  ~              \\ \hline

 ~$\sigma$~       &~ $\textrm{the activate function, e.g., Relu function}$ ~      \\ \hline
 ~$c_{t-1}$~        &~ $\textrm{the number of filters in layer t-1}$  ~                    \\ \hline

\end{tabular}
}\vspace{-10pt}
\end{table}

\begin{table*}
\centering {
\tiny
\scriptsize
\vspace{-0pt}
\caption{Information of the Graph Datasets}\label{T:GraphInformation} \vspace{0pt}
\begin{tabular}{|c||c||c||c||c||c||c||c||c|}
\hline
~Datasets ~        & ~MUTAG  ~  & ~PROTEINS~& ~D\&D~       & ~GatorBait~ & ~Reeb  ~   & ~IMDB-B~  & ~IMDB-M~   & ~RED-B~\\ \hline \hline
~Max \# vertices~  & ~$28$~     & ~$620$~   &  ~$5748$~    & ~$545$~     & ~$220$~    & ~$136$~   & ~$89$~     & ~$3783$~\\ \hline
~Mean \# vertices~ & ~$17.93$~  & ~$39.06$~ &  ~$284.30$~  & ~$348.70$~  & ~$95.42$~  & ~$19.77$~ & ~$13.00$~  & ~$429.61$~\\  \hline
~\# graphs~        & ~$188$~    & ~$1113$~  &  ~$1178$~    & ~$100$~     & ~$300$~    &  ~$1000$~ & ~$1500$~   & ~$2000$~    \\ \hline
~\# vertex labels~ & ~$7$~      & ~$61$~    &  ~$82$~      & ~$78$~       & ~$32$~      &  ~$-$~    & ~$-$~      & ~$-$~   \\ \hline
~\# classes~       & ~$2$~      & ~$2$~     &  ~$2$~       & ~$30$~      &  ~$20$~    & ~$2$~     &  ~$3$~     & ~$2$~   \\ \hline
~Description~      & ~Bioinformatics~ & ~Bioinformatics~&  ~Bioinformatics~  & ~Vision~    & ~Vision~   &  ~Social~ & ~Social~   & ~Social~   \\ \hline

\end{tabular}
} \vspace{-15pt}
\end{table*}

\textbf{Discussions:} Comparing to existing state-of-the-art graph convolution networks, the proposed Aligned Vertex Convolution Network (AVCN) model has a number of advantages.

\textbf{First}, unlike the Neural Graph Fingerprint Network (NGFN) model~\cite{DBLP:conf/nips/DuvenaudMABHAA15} and the Diffusion Convolution Neural Network (DCNN) model~\cite{DBLP:conf/nips/AtwoodT16} that both employ a SumPooling layer to directly sum up the extracted local-level vertex features from the convolution operation as global-level graph features. The proposed AVCN model focuses more on learning local structural features through the proposed aligned vertex grid structure. Specifically, Figure~\ref{f:vconv} indicates that the associated vertex convolution operation of the proposed AVCN model can convert the original aligned vertex grid structure into a new grid structure, by packing the aligned vertex features from the original grid structure into the new grid structure. Thus, \textbf{the new grid structure can be seen as a new extracted aligned vertex grid structure with a reduced number of aligned vertices}. As a result, the proposed AVCN model can gradually extract multi-scale local-level vertex features through a number of stacked vertex convolution layers, and encapsulate more significant local structural information than the NGFN and DCNN models associated with SumPooling.

\textbf{Second}, similar to the proposed AVCN model, both the PATCHY-SAN based Graph Convolution Neural Network (PSGCNN) model~\cite{DBLP:conf/icml/NiepertAK16} and the Deep Graph Convolution Neural Network model~\cite{DBLP:conf/aaai/ZhangCNC18} need to rearrange the vertex order of each graph structure and transform each graph into the fixed-sized vertex grid structure. Unfortunately, both the PSGCNN and the DGCNN models sort the vertices of each graph based on the local structural descriptor, ignoring consistent vertex correspondence information between different graphs. By contrast, the proposed AVCN model associates with a transitive vertex alignment procedure to transform each graph into an aligned fixed-sized vertex grid structure. As a result, only the proposed AVCN model can integrate the precise structural correspondence information over all graphs under investigations.

\textbf{Third}, when the PSGCNN model and the DGCNN model form fixed-sized vertex grid structures, some vertices with lower ranking will be discarded. This in turn leads to significant information loss. By contrast, the required aligned vertex grid structures for the proposed AVCN model can encapsulate all the original vertex features from the original graphs. As a result, the proposed AVCN overcomes the shortcoming of information loss arising in the PSGCNN and DGCNN models.


\section{Experiments}\label{s4}
In this section, we compare the performance of the proposed AVCN model to both state-of-the-art graph kernels and deep learning methods on graph classification problems on eight standard graph datasets. These datasets are abstracted from bioinformatics, computer vision and social networks. A selection of statistics of these datasets are shown in Table.\ref{T:GraphInformation}.

\textbf{Experimental Setup:} We evaluate the performance of the proposed AVCN model on graph classification problems against a) six alternative state-of-the-art graph kernels and b) six alternative state-of-the-art deep learning methods for graphs. Specifically, the graph kernels include 1) Jensen-Tsallis q-difference kernel (JTQK) with $q=2$~\cite{DBLP:conf/pkdd/Bai0BH14}, 2) the Weisfeiler-Lehman subtree kernel (WLSK)~\cite{shervashidze2010weisfeiler}, 3) the shortest path graph kernel (SPGK) \cite{DBLP:conf/icdm/BorgwardtK05}, 4) the shortest path kernel based on core variants (CORE SP)~\cite{DBLP:conf/ijcai/NikolentzosMLV18}, 5) the random walk graph kernel (RWGK)~\cite{DBLP:conf/icml/KashimaTI03}, and 6) the graphlet count kernel (GK)~\cite{DBLP:journals/jmlr/ShervashidzeVPMB09}. The deep learning methods include 1) the deep graph convolutional neural network (DGCNN)~\cite{DBLP:conf/aaai/ZhangCNC18}, 2) the PATCHY-SAN based convolutional neural network for graphs (PSGCNN)~\cite{DBLP:conf/icml/NiepertAK16}, 3) the diffusion convolutional neural network (DCNN)~\cite{DBLP:conf/nips/AtwoodT16}, 4) the deep graphlet kernel (DGK)~\cite{DBLP:conf/kdd/YanardagV15}, 5) the graph capsule convolutional neural network (GCCNN)~\cite{DBLP:journals/corr/abs-1805-08090}, and 6) the anonymous walk embeddings based on feature driven (AWE)~\cite{DBLP:conf/icml/IvanovB18}.

\begin{figure}
 \vspace{-0pt}
 \centering
\includegraphics[width=0.85\linewidth]{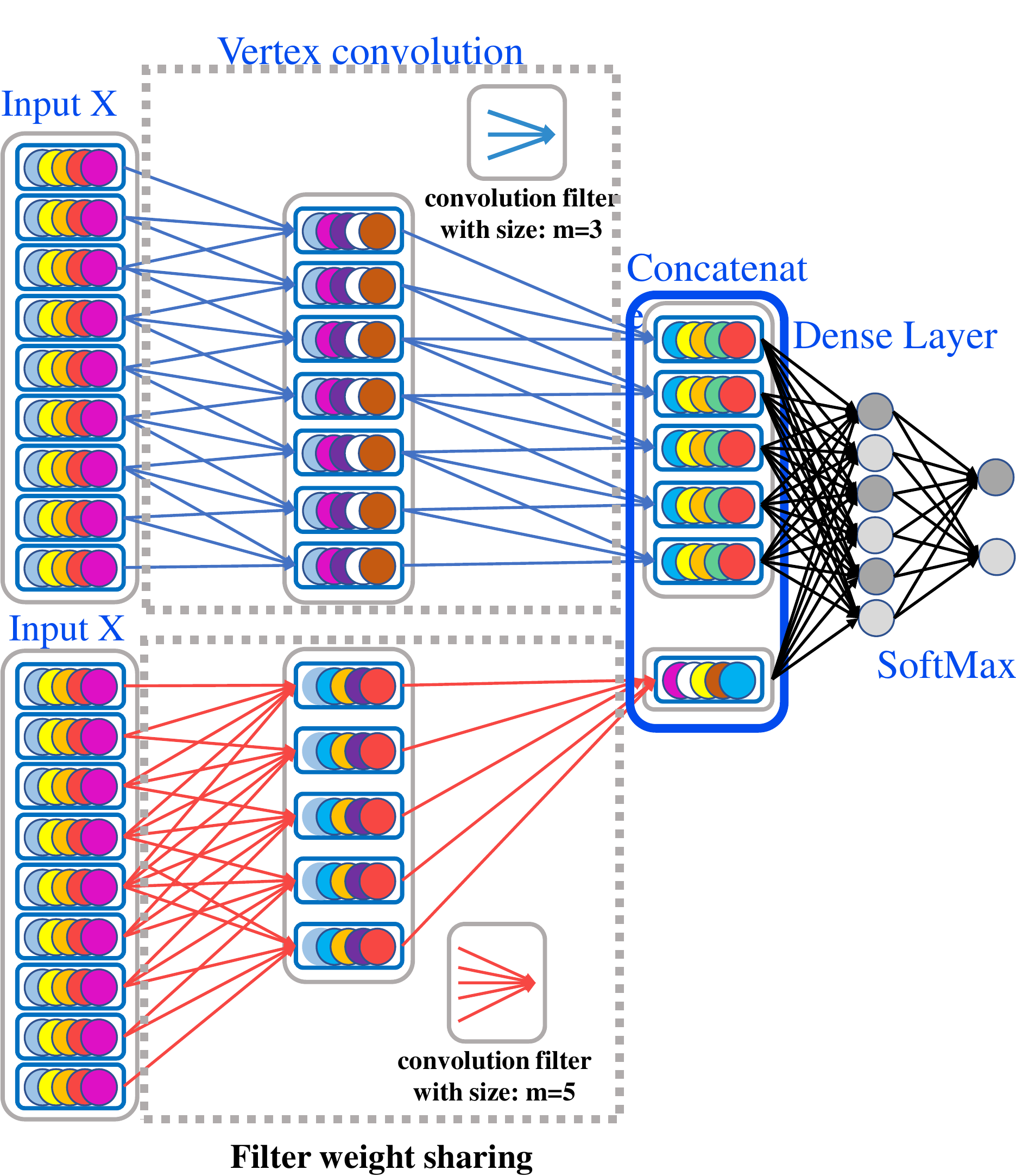}
 \vspace{-10pt}
  \caption{An example of the ACVN architecture.}\label{f:vcn_arc}
 \vspace{-22pt}
\end{figure}

\begin{table*}
\centering {
\tiny
\scriptsize
\caption{Classification Accuracy (In $\%$ $\pm$ Standard Error) for Comparisons with Graph Kernels.}\label{T:ClassificationGK}
\vspace{0pt}
\begin{tabular}{|c|c|c|c|c|c|c|c|c|c|}

  \hline
 ~Datasets~& ~MUTAG  ~         & ~PROTEINS~       & ~D\&D~              & ~GatorBait~      & ~Reeb~           & ~IBDM-B~        & ~IBDM-M~         & ~RED-B~\\ \hline \hline

 ~\textbf{AVCN}~  & ~$\textbf{89.06}\pm0.90$~  & ~$\textbf{75.71}\pm0.65$~ &  ~$\textbf{80.90}\pm0.97$~   & ~$\textbf{19.00}\pm.75$~ & ~$\textbf{67.00}\pm0.91$ ~& ~$\textbf{73.20}\pm.90$  & ~$51.14\pm.87$   & ~$\textbf{91.00}\pm0.20$\\ \hline

  ~JTQK~   & ~$85.50\pm0.55$~  & ~$72.86\pm0.41$~ &  ~$79.89\pm0.32$~   & ~$11.40\pm0.52$~ &~$60.56\pm0.35$~  &~$72.45\pm0.81$~ &  ~$50.33\pm0.49$~ & ~$77.60\pm0.35$\\ \hline

  ~WLSK~   & ~$82.88\pm0.57$~  & ~$73.52\pm0.43$~ &  ~$79.78\pm0.36$~   & ~$10.10\pm0.61$~ &~$58.53\pm0.53$~  &~$71.88\pm0.77$~ &  ~$49.50\pm0.49$~ & ~$76.56\pm0.30$\\   \hline

  ~SPGK~   & ~$83.38\pm0.81$~  & ~$75.10\pm0.50$~ &  ~$78.45\pm0.26$~   & ~$9.00\pm0.75$~ &~$55.73\pm0.44$~   &~$71.26\pm1.04$~ &  ~$\textbf{51.33}\pm0.57$~ & ~$84.20\pm0.70$\\  \hline

  ~CORE SP~& ~$88.29\pm1.55$~  & ~$-$~            &  ~$77.30\pm0.80$~   & ~$-$~           &~$-$~             &~$72.62\pm0.59$~ &  ~$49.43\pm0.42$~ & ~$90.84\pm0.14$\\  \hline

  ~  GK~   & ~$81.66\pm2.11$~  & ~$71.67\pm0.55$~ &  ~$78.45\pm0.26$~   & ~$8.40\pm.83$~ &~$22.96\pm0.65$~  &~$65.87\pm0.98$~ &  ~$45.42\pm0.87$~ & ~$77.34\pm0.18$\\ \hline

  ~RWGK~   & ~$80.77\pm0.72$~  &~$74.20\pm0.40$~  &  ~$71.70\pm0.47$~   & ~$7.00\pm0.77 $~&~$32.47\pm0.69$~             &~$67.94\pm0.77$~ &  ~$46.72\pm0.30$~ & ~$72.73\pm0.39$           \\ \hline

\end{tabular}
%
%
%
%
%
%
%
%
%
} \vspace{-15pt}
\end{table*}
\begin{table*}
\centering {
 \tiny
\scriptsize
\caption{Classification Accuracy (In $\%$ $\pm$ Standard Error) for Comparisons with Graph Convolutional Neural Networks.}\label{T:ClassificationGCNN}
\vspace{0pt}

\begin{tabular}{|c|c|c|c|c|c|c|c|}

  \hline
 ~Datasets~& ~MUTAG  ~       & ~PROTEINS~      & ~D\&D~            & ~IBDM-B~        & ~IBDM-M~         & ~RED-B~    \\ \hline \hline

 ~\textbf{AVCN}~  & ~$\textbf{89.06}\pm0.90$~& ~$75.71\pm0.65$~& ~$\textbf{80.90}\pm0.97$~ & ~$\textbf{73.20}\pm.90$  & ~$51.14\pm.87$   & ~$\textbf{91.00}\pm0.20$\\ \hline

  ~DGCNN~  & ~$85.83\pm1.66$~& ~$75.54\pm0.94$~& ~$79.37\pm0.94$~ & ~$70.03\pm0.86$ & ~$47.83\pm0.85$  & ~$76.02\pm1.73$\\ \hline

  ~PSGCNN~ & ~$88.95\pm4.37$~& ~$75.00\pm2.51$~& ~$76.27\pm2.64$~ & ~$71.00\pm2.29$ & ~$45.23\pm2.84$  & ~$86.30\pm1.58$\\ \hline

  ~DCNN~   & ~$66.98$~       & ~$61.29\pm1.60$~& ~$58.09\pm0.53$~ & ~$49.06\pm1.37$ & ~$33.49\pm1.42$  & ~$-$\\ \hline

  ~GCCNN~& ~$-$~             & ~$\textbf{76.40}\pm4.71$~& ~$77.62\pm4.99$~  & ~$71.69\pm3.40$ & ~$48.50\pm4.10$  & ~$87.61\pm2.51$\\ \hline

  ~DGK~   & ~$82.66\pm1.45$~ & ~$71.68\pm0.50$~& ~$78.50\pm0.22$~  & ~$66.96\pm0.56$ & ~$44.55\pm0.52$  & ~$78.30\pm0.30$\\ \hline

  ~AWE~& ~$87.87\pm9.76$~    & ~$-$~           & ~$71.51\pm4.02$~  & ~$73.13\pm3.28$ & ~$\textbf{51.58}\pm4.66$  & ~$82.97\pm2.86$\\ \hline

\end{tabular}
} \vspace{-10pt}
\end{table*}

For the experiment, \textbf{the proposed AVCN model uses the same network structure on all graph datasets}. Specifically, we set the channel of each vertex convolution operation as $32$, and the number of the prototype representations as $M=64$, i.e., the vertex numbers of the aligned vertex grid structures for the graphs in any dataset are all $64$. To extract different hierarchical multi-scale local vertex features, we propose to input the aligned vertex grid structure of each graph to a family of paralleling stacked vertex convolution layers associated with different convolution filter sizes. Specifically, the architecture of the AVCN model is $C_{64}^{4:(3;5;7;9)}$-$C_{64}^{4:(3;5;7;9)}$-$C_{64}^{4:(3;5;7;9)}$-$F_{64}$. Here, $C_k^{f:(i,j,x,y)}$ denotes a vertex convolution layer consisting of $f$ paralleling vertex convolution filters each with $k$ channels, and the filter sizes are $i$, $j$, $x$ and $y$ respectively. $F_k$ denotes a fully-connected layer consisting of $k$ hidden units. An example of the architecture $C_5^{2:(3;5)}$-$C_5^{2:(3;5)}$-$F6$ for the proposed AVCN model are shown in Figure~\ref{f:vcn_arc}. We set the stride of each filter in $C_k^{f:(i,j,x,y)}$ layer as $1$. With extracted patterns learned from the paralleling stacked vertex convolution layers to hand, we concatenate them and add a new fully-connected layer followed by a Softmax layer to learn the graph class. We set the dropout rate for the fully connected layer as $0.5$. We employ the rectified linear units (ReLU) as the active function for the convolution layers. The only hyperparameter that we need to be optimized is the learning rate, the number of epochs, and the batch size for the mini-batch gradient decent algorithm. To optimize the our AVCN model, we utilize the Stochastic Gradient Descent with the Adam updating rules. Finally, note that, our AVCN model needs to construct the prototype representations to identify the transitive vertex alignment information over all graphs. In this evaluation we proposed to compute the prototype representations from both the training and testing graphs. Thus, our model is an instance of transductive learning~\cite{DBLP:conf/uai/GammermanAV98}, where all graphs are used to compute the prototype representations but the class labels of the testing graphs are not used during the training process. For our model, we perform $10$-fold cross-validation to compute the classification accuracies, with nine folds for training and one fold for testing. For each dataset, we repeat the experiment 10 times and report the average classification accuracies and standard errors in Table.\ref{T:ClassificationGK}.


For the alternative kernel methods, we set the parameters of the maximum subtree height for both the WLSK and JTQK kernels as $10$, based on the previous empirical studies in the original papers. For each alternative graph kernel, we perform $10$-fold cross-validation associated with the LIBSVM implementation of C-Support Vector Machines (C-SVM) to compute the classification accuracies. We repeat the experiment 10 times for each kernel and dataset and we report the average classification accuracies and standard errors in Table.\ref{T:ClassificationGK}. Note that for some kernels we directly report the best results from the original corresponding papers, since the evaluation of these kernels followed the same setting of ours. On the other hand, for the alternative deep learning methods, we report the best results for the PSGCNN and DGK models from their original papers. Note that, these methods were evaluated based on the same setting with the proposed AVCN model. For the DCNN model, we report the best results from the work of Zhang et al.,~\cite{DBLP:conf/aaai/ZhangCNC18}, following the same setting of ours. For the AWE model, we report the classification accuracies of the feature-driven AWE, since the author have stated that this kind of AWE model can achieve competitive performance on label dataset. Finally, note that the PSGCNN model can leverage additional edge features, most of the graph datasets and the alternative methods do not leverage edge features. Thus, we do not report the results associated with edge features in the evaluation. The classification accuracies and standard errors for each deep learning method are shown in Table.\ref{T:ClassificationGCNN}. Note that, the alternative deep learning methods have been evaluated on the Reeb and GatorBait datasets abstracted from computer vision by any author, we do not include the accuracies for these methods.

\textbf{Experimental Results and Discussions:} Table.\ref{T:ClassificationGK} and Table.\ref{T:ClassificationGCNN} indicate that the proposed AVCN model can outperform the alternative state-of-the-art methods including either the graph kernels or the deep learning methods for graphs. Specifically, for the alternative graph kernels, only the accuracy of the SPGK kernel on the IBDM-M dataset is a little higher than that of the proposed AVCN model. On the other hand, for the alternative deep learning methods, only the accuracies of the GCCNN model on the PROTEINS dataset and the AWE model on the IMDB-M dataset are a little higher than those of the proposed AVCN model. The reasons for the effectiveness are threefold. First, these alternative graph kernels are typical examples of R-convolution kernels and are based on measuring any pair of substructures, ignoring the correspondence information between the substructures. By contrast, the proposed model associated with aligned vertex grid structure incorporates the transitive alignment information between graphs, and thus better reflect graph characteristics. Furthermore, the C-SVM classifier associated with graph kernels can only be seen as a shallow learning framework~\cite{DBLP:conf/icassp/ZhangLYG15}. By contrast, the proposed model can provide an end-to-end deep learning architecture, and thus better learn graph characteristics. Second, similar to the alternative graph kernels, all the alternative deep learning methods also cannot integrate the correspondence information between graphs into the learning architecture. Especially, the PSGCNN and DGCNN models need to reorder the vertices and some vertices may be discarded, leading to information loss. By contrast, the associated aligned vertex grid structures can preserve all the information of original graphs. Third, unlike the proposed model, the DCNN model needs to sum up the extracted local-level vertex features as global-level graph features. By contrast, the proposed model can learn richer multi-scale local-level vertex features. The experiments demonstrate the effectiveness of the proposed model.

\section{Conclusion}\label{s6}
In this paper, we have developed a new aligned vertex convolutional network model for graph classification. The proposed model cannot only integrates the precise structural correspondence information between graphs but also minimises the loss of structural information residing on local-level vertices. Experiments demonstrate the effectiveness of the proposed vertex convolution network model.

\bibliographystyle{named}
\bibliography{ijcai19}

\end{document}